\documentclass[journal]{IEEEtran}
\usepackage{times}
\usepackage{epsfig}
\usepackage{graphicx}
\usepackage{amsmath}
\usepackage{amssymb}
\usepackage{algorithmic}
\usepackage{algorithm}
\usepackage{subfig}
\usepackage{array}
\usepackage{enumerate}
\usepackage{color}
\usepackage[switch]{lineno}
\usepackage{tabularx}
\usepackage{booktabs}

\ifCLASSINFOpdf
\else
\fi
\begin{document}
%
\title{Content-Adaptive Sketch Portrait Generation by Decompositional Representation Learning}
%
%
%

\author{Dongyu Zhang, Liang Lin, Tianshui Chen, Xian Wu, Wenwei Tan, and Ebroul Izquierdo 
\thanks{This work is partially supported by the State Key Development Program under Grant No. 2016YFB1001000, the National Natural Science Foundation of China under Grant No. 61401125, 61671182, and Science and Technology Program of Guangdong Province under Grant No. 2015B010128009 and. This work is also supported by the fund from Huawei Technologies Co., Ltd. (Corresponding author: Liang Lin)

D. Zhang, L. Lin, T. Chen, X. Wu are with the School of Data and Computer Science, Sun Yat-sen University, Guangzhou, China. (Email: cszhangdy@163.com; linliang@ieee.org; tianshuichen@gmail.com; sysuwuxian@gmail.com).

W. Tan is with the Hisilicon Technologies co., LTD. (Email: tanwenwei@hisilicon.com).

Ebroul Izquierdo is with the School of Electronic Engineering and Computer Science, Queen Mary University of London, London, U.K. (Email: ebroul.izquierdo@qmul.ac.uk )}
}

%
%

\markboth{IEEE TRANSACTIONS ON IMAGE PROCESSING, VOL. 26, NO. 1, JANUARY, 2017}%
{Dongyu Zhang \MakeLowercase{\textit{et al.}}: Content-Adaptive Sketch Portrait Generation by Decompositional Representation Learning}
%



\maketitle

\begin{abstract}
Sketch portrait generation benefits a wide range of applications such as digital entertainment and law enforcement. Although plenty of efforts have been dedicated to this task, several issues still remain unsolved for generating vivid and detail-preserving personal sketch portraits. For example, quite a few artifacts may exist in synthesizing hairpins and glasses, and textural details may be lost in the regions of hair or mustache. Moreover, the generalization ability of current systems is somewhat limited since they usually require elaborately collecting a dictionary of examples or carefully tuning features/components. In this paper, we present a novel representation learning framework that generates an end-to-end photo-sketch mapping through structure and texture decomposition. In the training stage, we first decompose the input face photo into different components according to their representational contents (i.e., structural and textural parts) by using a pre-trained Convolutional Neural Network (CNN). Then, we utilize a Branched Fully Convolutional Neural Network (BFCN) for learning structural and textural representations, respectively. In addition, we design a Sorted Matching Mean Square Error (SM-MSE) metric to measure texture patterns in the loss function. In the stage of sketch rendering, our approach automatically generates structural and textural representations for the input photo and produces the final result via a probabilistic fusion scheme. Extensive experiments on several challenging benchmarks suggest that our approach outperforms example-based synthesis algorithms in terms of both perceptual and objective metrics. In addition, the proposed method also has better generalization ability across dataset without additional training.
\end{abstract}


%
\IEEEpeerreviewmaketitle

\section{Introduction}
%
%
%
%
\IEEEPARstart{S}{ketch} portrait generation has widespread utility in many applications \cite{wang2009face, song_eccv14_sketch, zhang2015face}. For example, in the law enforcement, when it is impossible to get the photo of criminal, a sketch portrait drawn based on the description of eyewitness may help the policemen to quickly identify the suspect by utilizing automatical sketch-based retrieval in the mug-shot database. In digital entertainment, people like to render their photos into sketch style and use them as the avatars on social media for enjoyment.

\begin{figure}[thbp]
\centering
\includegraphics[width=0.48\textwidth]{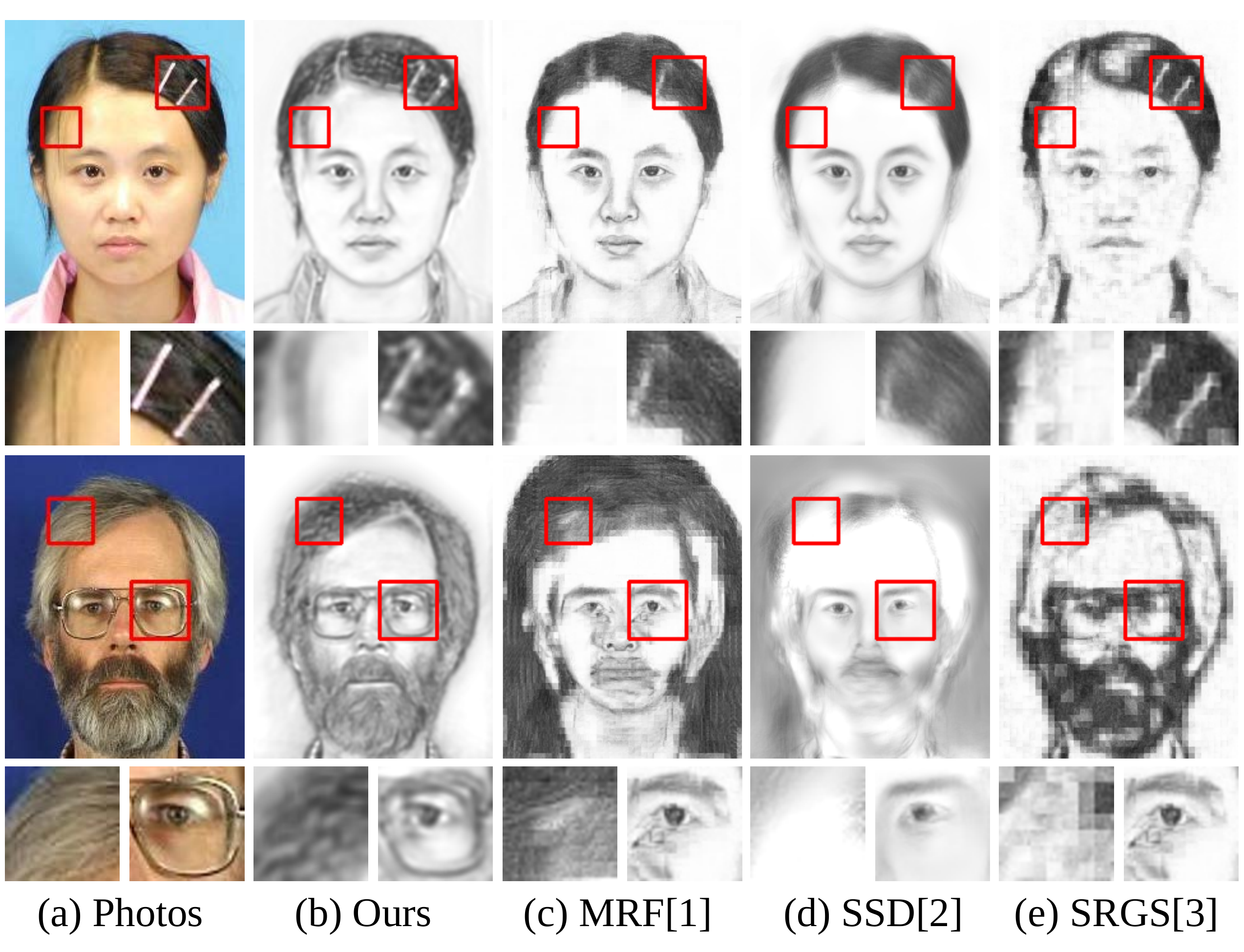}
\caption{Illustration results of existing methods and the proposed approach.}
\label{fig:motivation}
\end{figure}

Despite the widespread applications of sketch portrait, it remains a challenging problem to generate vivid and detail-preserved sketch because of the great difference between photo and sketch. To the best of our knowledge, most of existing approaches generate sketch portraits based on the synthesis of training examples.
Given a photo patch, these methods find similar patches in the training set and use their corresponding sketch patches to synthesize the sketch of input photo. Although impressive results have been received, there remains several issues in these methods. As shown in Fig. \ref{fig:motivation}, the synthesis results of non-facial factors of these example-based methods are not satisfied, such as hairpins and glasses \cite{wang2009face, zhang2015face}. Because of the great variations in appearance and geometry of these decorations, it is easy to involve artifacts in the synthesis results. Besides some methods \cite{song_eccv14_sketch, zhang2015face} average the candidate sketches to generate smoothed results. They may produce acceptable sketches for face part, but always fail to preserve textural details, such as the hair region. Finally, the performance of these example-based methods are only acceptable when training and test samples originate from the same dataset, however, this situation is rarely happened in practice.

Aiming at alleviating the aforementioned problems, we propose to learn sketch representations directly from raw pixels of input photos, and develop a decompositional representation learning framework to generate an end-to-end photo-sketch mapping through structure and textural decomposition. Given an input photo, our method first roughly decompose it into different regions according to their representational contents, such as face, hair and background. Then we learn structural representation and textural representation from different parts respectively. The structural representation learning mainly focuses on the facial part, while the textural representation learning mainly targets on preserving the fine-grained details of hair regions. Finally, the two representations are fused to generate the final sketch portrait via a probabilistic method.

\begin{figure*}[!thbp]
\centering
\includegraphics[width=0.9\textwidth]{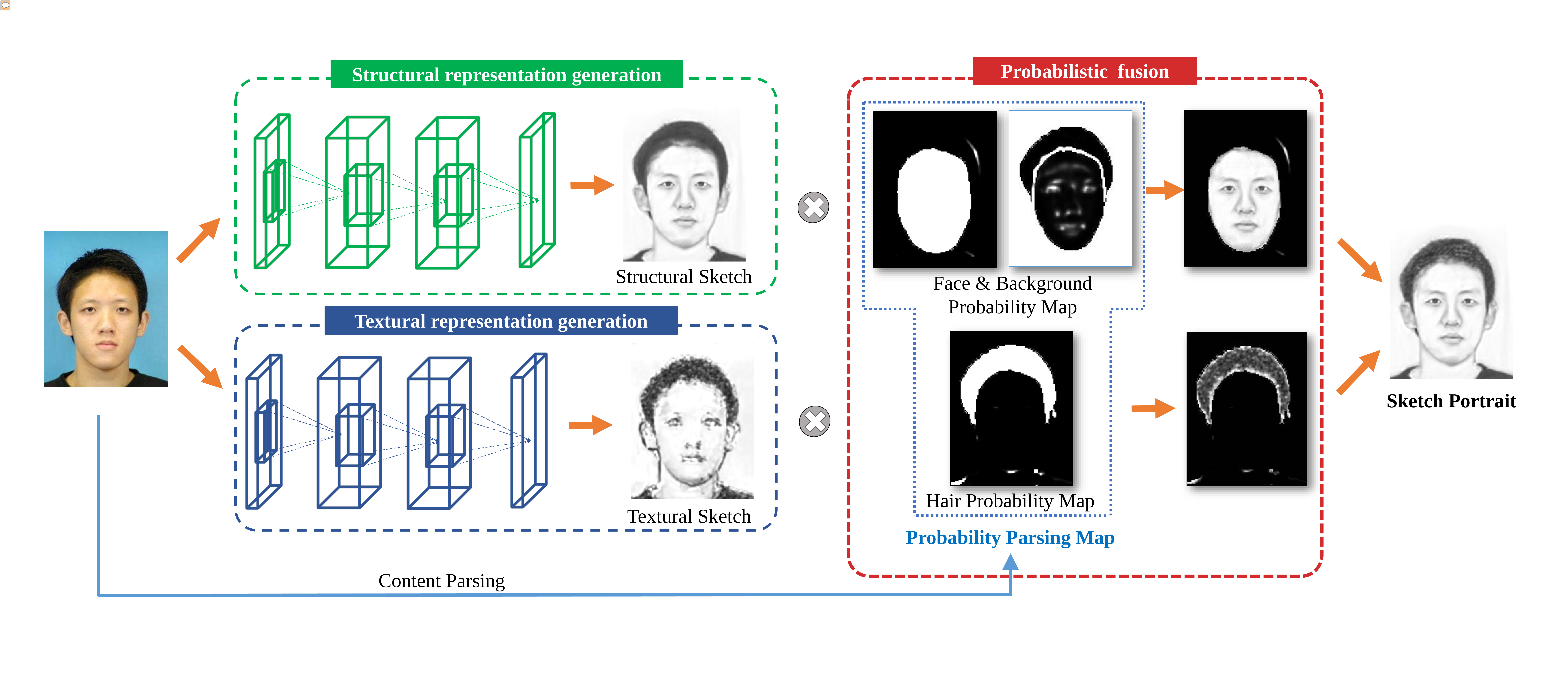}
\caption{Illustration of the pipeline of sketch portraits generation via the proposed framework. Our approach feeds an input photo into the branched fully convolutional network to produce a structural sketch and a textural sketch, respectively. Guided by the parsing maps, the two sketches are fused to get the final result via a probability fusion method.}
\label{fig:pipeline}
\end{figure*}

Specifically, in the training stage, we first adopt a pre-trained parsing network (P-Net) to automatically output a probability parsing map, which assigns a three-dimensional vector to each pixel of input photo to indicate its probability belonging to face, hair, and background. With the probability parsing map we can easily obtain the face regions and hair regions.
We then utilize a branched fully convolutional network (BFCN), which includes a structural branch and a textural branch, to learn the structural representation and textural representation respectively. We select patches of face part when training the structural branch and adopt mean square error (MSE) as its objective function.

For the textural branch, we feed it with patches selected from hair regions. As to the loss function of textural branch, we do not use MSE which is used in the training of structural branch. The reason is that different from structural regions, textural regions usually possess periodic and oscillatory natures \cite{bertalmio2003simultaneous, aujol2006structure, zhang2016robust}, and a point-to-point matching, such as MSE, is not effective enough to measure the similarity of two similar textural regions. Thus, directly applying MSE for textural branch learning can not well preserve the fine-grained textural details. To solve this problem, we propose a sorted matching mean square error (SM-MSE) for the training of textural branch of BFCN. SM-MSE can be regarded as applying an ascending sort operator before calculating MSE. Compared with MSE, it can effectively evaluate the similarity of two textural patterns. The detail of SM-MSE is described in Section III.

In the testing stage, given an input photo, we first use BFCN to learn its structural representation and textural representation. Then, the two representations are fused to generate final sketch portrait guided by the probability parsing maps. The pipeline of generating sketch portraits via BFCN is illustrated in Fig. \ref{fig:pipeline}.

The key contribution of this work is a task-driven deep learning method that achieves a new state-of-the-art performance for personal sketch portrait generation. Our framework is capable of learning the photo-sketch mapping in an end-to-end way, unlike the traditional approaches that usually require elaborately collecting a dictionary of examples or carefully tuning features/components. Moreover, the proposed SM-MSE metric is very effective to measure texture patterns during the representation learning, improving the expression of sketch portraits through capturing textural details.

The remainder of this paper is organized as follows. Section II reviews related works about sketch synthesis and convolutional neural networks. Section III describes the proposed decompositional representation learning framework for sketch portrait generation in detail. Extensive experimental results are provided in Section IV. Finally, Section V concludes this paper.

\section{Related work}
In this section, we first review the example-based sketch synthesis methods proposed in previous work. Then, we discuss different strategies which produce dense sketch outputs via neural networks.

\subsection{Sketch Portrait Generation via Synthesis-by-Exemplar}
Most works in sketch portrait generation focus on two kinds of sketches, namely profile sketches \cite{xu2008hierarchical} and shading sketches \cite{tang2004face}. Compared with the former, the shading sketches can not only use lines to reflect the overall profiles, but also capture the textural parts via shading. Thus, the shading sketches are more challenge to be modeled. We mainly study the automatic generation of shading sketches in this paper.

In most previous works, sketch portrait generation is usually modeled as a synthesis problems with assumption that similar photo images have similar sketch images. Tang and Wang \cite{tang2004face} proposed a sketch portrait generation method based on eigen transformation (ET). For each test photo image, this method searches similar photo images in a prepared training set, and then uses the corresponding sketch images to synthesize the sketch. The photo-to-sketch mapping is approximated as linear transform in ET-based method. However, this assumption may be too strong, especially when the hair regions are included. Liu et al. \cite{liu2005nonlinear} proposed a nonlinear method using locally linear embedding (LLE), which partitions the image into several overlapping patches and synthesizes each of these patches separately.
Recent works also partition the images into patches for further synthesizing. To fulfill the smoothness requirement between neighboring patches, Wang and Tang proposed a multiscale Markov Random Fields (MRF) model \cite{wang2009face}. But it is too computationally intensive to be applied in realtime situations. To reduce the synthesized artifacts, Song et al. \cite{song_eccv14_sketch} improved the LLE-based method \cite{liu2005nonlinear} by considering synthesis as an image denoising processing. However, the  high-frequency information is suppressed in their results. To enhance the generalization ability, Zhang et al. \cite{zhang2015face} designed a method called sparse representation-based greedy search (SRGS), which searches candidates globally under a time constraint. However, their results are inferior in preserving clear structures.

Several methods add a refinement step to recover vital details of the input photo to improve the visual quality and face recognition performance. Zhang et al. \cite{zhang2011svr} applied a support vector regression (SVR) based model to synthesize the high-frequency information. Similarly, Gao et al. \cite{gao2012face} proposed a method called SNS-SRE with two steps, i.e., sparse neighbor selection (SNS) to get an initial estimation and sparse representation based enhancement (SRE) for further improvement. Nevertheless, these post processing steps may brought in side effects, e.g., the results of SNS-SRE are out of sketch styles and become more likely to be natural gray level images.

\subsection{Dense Predictions via Convolutional Neural Networks}
The convolutional neural network (CNN) has been widely used in computer vision. Its typical structure contained a series of convolutional layers, pooling layers and full connected layers. Recently, CNN has achieved great success in large scale object localization \cite{sermanet-iclr-14, chen2016disc}, detection \cite{erhan2013scalable}, recognition \cite{lin2015deep,he2014spatial,lin2016cross,zhang2015bit} and classification \cite{krizhevsky2012imagenet, zeiler2014visualizing}.

Researchers also adopted CNNs to produce dense predictions. An intuitive strategy is to attach the output maps to the topmost layer for directly learning a global predictions. For examples, Wang et al. \cite{wang2014deep} adopted these strategy for generic object extraction, and Luo et al. \cite{luo2013pedestrian} applied a similar configuration for pedestrian parsing. Nevertheless, this strategy often produces coarse outputs, since the parameters in networks grow dramatically when enlarging the output maps. To produce finer outputs, Eigen et al. \cite{sermanet-iclr-14} applied another network which refined coarse predictions via information from local patches in the depth prediction task. A similar idea was also proposed by Wang et al. \cite{wang2014designing}, which separately learns global and local processes and uses a fusion network to fuse them into the final estimation of the surface normal. Surprisingly, the global information can be omitted in some situations, e.g., Dong et al. \cite{dong2014learn, dong2014image} applied a CNN only included three convolutional layers for image super resolution. Though this network has a small receptive field and is trained on local patch samples, it works well for the strict alignment of samples in this specific task.

\begin{figure*} [!thbp] \centering
\includegraphics[width=0.9\textwidth]{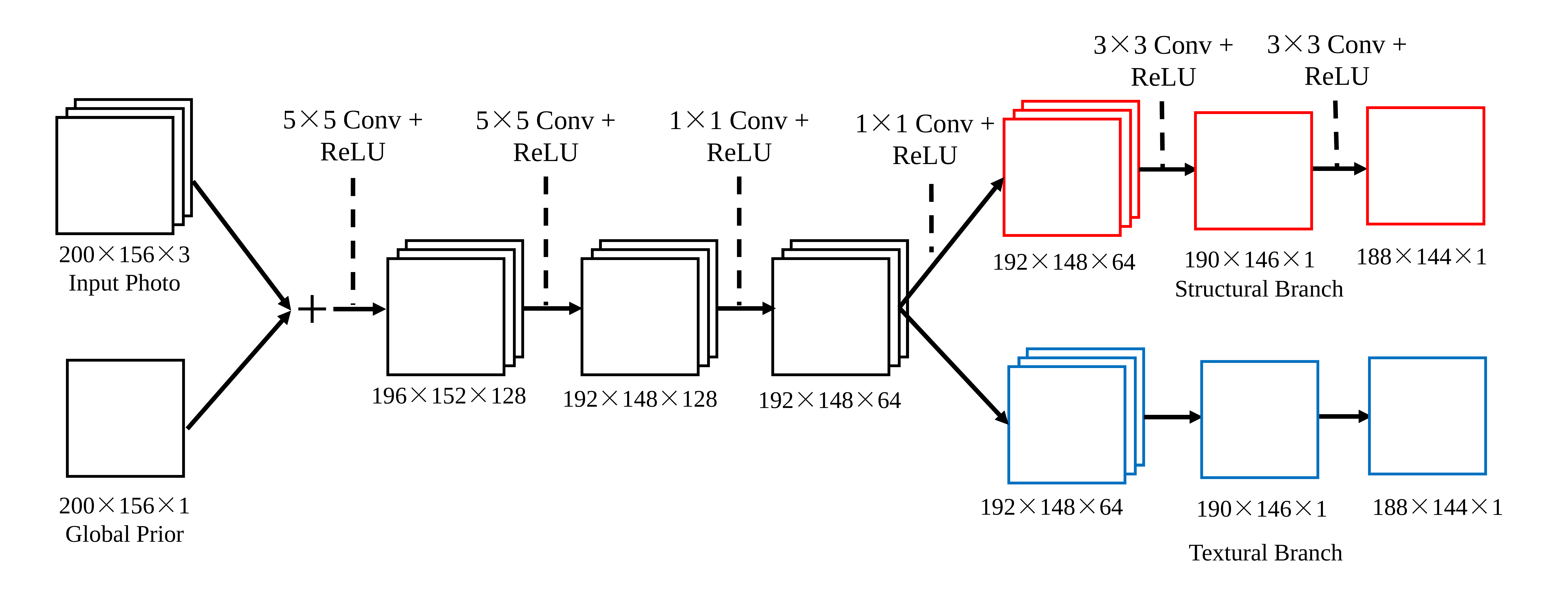}

\caption{The architecture of Branched Fully Convoluational Neural Network. A photo and global prior are taken as the input. They are fed into three shared convolutional layers with the kernel sizes $(5\times5)$, $(5\times5)$ and $(1\times1)$, and then they pass through two branches with additional three convolutional layers with the kernel sizes $(1\times1)$, $(3\times3)$ and $(3\times3)$. The two output layers are connected with specific objective functions for predictions of structures and textures, respectively. }
\label{fig:G-Net}
\end{figure*}

\section{Sketch generation via Decompositional Representation Learning}

In this paper, we propose a representation learning framework for an end-to-end photo-sketch mapping via structure and texture decomposition.
Given an image, it can be decomposed into structural components and textural components \cite{le2014cartoon+}. The geometric and smoothly-varying component, referred to as structural component or cartoon, is composed of object hues and boundaries, while the texture is an oscillatory component capturing details and noise. Thus, in the proposed framework, we separately learns the structural and textural representations of photo portrait.

In the training stage, by using a probability parsing map, a photo is automatically decomposed into different semantic parts, i.e., face, hair, and background. Then, we utilize a branched fully convolutional network (BFCN) to learn the structural and textural representation respectively. Patches from face region are fed to BFCN to train the structural branch, while patches from hair region are fed into BFCN to train its textural branch, respectively. In the test stage, given a test photo, BFCN automatically learns a structure-preserved sketch and a texture-preserved sketch, which are further fused to generate the final sketch portrait via a probabilistic method.

In the following, we will first introduce the probability parsing map, and then describe the architecture and the specific training strategy of BFCN. The probabilistic fusion method is presented at the end of this section.

\subsection{Probability Parsing map}

Inspired by previous works \cite{long_shelhamer_fcn, liu2015cvpr}, we design a fully convolutional network pre-trained on Helen dataset to automatically parse a face photo into semantic regions of face, hair and background. This network is called parsing net (P-Net), which consists of eight convolutional layers with ReLUs as activation functions. The first three convolutional layers are followed by pooling layers and local response normalization layers \cite{krizhevsky2012imagenet}. An average probability map of the face, hair, and background, is also adopted as nonparametric priors to provide a global regularization. In the inference stage, we feed this network with the full-size $(200\times156)$ photo. Then P-Net generates three maps of the size $(100\times78)$, corresponding to the probability distributions of face, hair and background of pixels in the photo respectively.

We adopt a softmax classifier on the top of P-Net to learn the probabilistic parsing probability maps. For an input image $\mathbf{X}$, we use $\mathbf{Y}$ to denote its ground truth probability parsing map. For each pixel $y \in \mathbf{Y}$, and its receptive field is denoted as $\mathbf{x}$. Let $\mathbf{w}_p$ denote the parameters of P-Net. Then the topmost output of P-Net can be denoted as $h=f(\mathbf{x},\mathbf{w}_p)$.

Thus the predictions of softmax classifier can be formulated as
\begin{equation}
\mathbf{P}(y=l|h,\mathbf{w})=\frac{exp \left( (\mathbf{w}^l)^{\mathrm T}h \right)}{\sum_{l=1}^{3}{exp\left((\mathbf{w}^l)^{\mathrm T}h\right)}},
\label{equation:softmax}
\end{equation}
where $l=\{1,2,3\}$ indicating the class labels of $y$, i.e., face, hair and background, $\mathbf{w}$ denotes the weight of softmax classifier, and $\mathbf{w}^l$ denotes the weight for the $l$-th class. Thus, for a single image $\mathbf{X}$ and its corresponding probability parsing map $\mathbf{Y}$, we can formulate the objective of P-Net as
\begin{equation}
\mathcal{L}_p(\mathbf{X}, \mathbf{Y}, \mathbf{w}_p, \mathbf{w}) \\
 = -\frac{1}{|\mathbf{Y}|}\sum_{y \in \mathbf{Y}}\sum_{l=1}^{3}{\mathbf{l}(y=l)\log P(y=l|h,\mathbf{w})},
\label{equation:softmax-loss}
\end{equation}
where $\mathbf{l}(\cdot)$ is the indicator function.

\subsection{Branched Fully Convolutional Network}

We utilize a branched fully convolutional neural network, i.e., BFCN, to learn the structural and textural representations of photo portrait respectively. The architecture of BFCN is shown in Fig. \ref{fig:G-Net}. BFCN consists of six convolutional layers of rectified linear functions (ReLUs \cite{nair2010rectified}) as the activation functions. We share the features of first three layers in BFCN for computational efficiency, and adopt two sibling output layers to produce the structural and textural predictions. As the receptive field of BFCN is small, it may fail to predict satisfactory results via small local information. Thus we add a nonparametric prior to provide a global regularization as introduced in previous work \cite{liu2015cvpr}. More precisely, we average of all the aligned ground truth sketches to get an average sketch portrait and attach it after color channels as the network input. Though we only feed BFCN with patches in the training stage, this network can be fed with full size images in the testing time due to the translation invariance of the convolutional operator.

There are two sibling branches in BFCN, i.e., structural branch and textural branch. In the training stage, patches from face part are fed to structural branch to learn the structural representations, while patches from hair region are fed into textural branch for textural representation learning. We adopt different objective functions to train the two branches. Let  $\mathcal{L}_g$ denotes the total objective function of BFCN. Then, $\mathcal{L}_{g}$ can be formulated as
\begin{equation}
\mathcal{L}_g = \mathcal{L}_{s} + \alpha \mathcal{L}_{t},
\label{equation:G-Net-objective}
\end{equation}

\noindent where $\mathcal{L}_{s}$ denotes the structural objective function, $\mathcal{L}_{t}$ denotes the textural objective function, and $\alpha$ is a scaling factor to balance the two objective function terms. In the following, we describe the definition of $\mathcal{L}_{s}$ and $\mathcal{L}_{t}$ and the training strategies respectively.

\subsubsection{Structural branch training}

Patches from the face regions are fed to BFCN for the structural representation, and we apply MSE as the objective function of structural branch. Let $(p_s,s_s)$ denote a structural training patch pair, and  $\mathbf{w}_g$ and $\mathbf{w}_s$ denote the parameters in the shared layers and the structural branch. The structural objective function $\mathcal{L}_{s}$ can be formulated as

\begin{equation}
\mathcal{L}_{s} = \frac{1}{|\mathbf{P}_s|}\sum_{p_s \in \mathbf{P}_s} \mathrm{MSE}(\hat{s}_s, s_s),
\label{equation:structrual-generation-objective}
\end{equation}
where $\hat{s}_{s}=f(p_s, \mathbf{w}_g, \mathbf{w}_{s})$ denotes the structural prediction of $s_s$, and $|\mathbf{P}_s|$ denotes the total number of training photo patch set $\mathbf{P}_s$. The $\mathrm{MSE}(\cdot)$ in Eq. (\ref{equation:structrual-generation-objective}) can be formulated as

\begin{equation}
\mathrm{MSE}(\hat{s}_s, s_s) = \frac{1}{|s_s|}\sum_{s_s^i \in s_s} \left ( \hat{s}_{s}^i - s_s^i \right ) ^ 2 ,
\label{equation:MSE}
\end{equation}
where $s_s^i$ denotes the  $i$-th ground truth pixel of a structural sketch patch $s_s$, and $\hat{s}_s^i \in \hat{s}_s$ denotes the corresponding prediction.

In the training set, each photo and its corresponding sketch are cropped into small patches in the same size to form the training photo-sketch patch pairs. However, as the photo and its corresponding sketch are only roughly aligned by facial landmarks, there are a lot of structurally unaligned patch pairs \cite{wang2009face}. Those unaligned patch pairs will greatly degrade the visual quality of final results. Thus, it is necessary to filter them before structural representation learning.

We assume that a photo patch and a sketch are aligned if they have high structural similarity. To measure their structural similarity, we first utilize the Sobel operator to exact the edge maps of two patches, and then adopt the Structural Similarity (SSIM) \cite{wang2004image} index to evaluate the similarity between the two edge maps. Then, we filter out the patch pairs with SSIM indexes lower than a threshold (e.g., $\le 0.6$ in this paper).

\subsubsection{Textural branch training}
\begin{figure}[thbp] \centering

\includegraphics[width=0.45\textwidth]{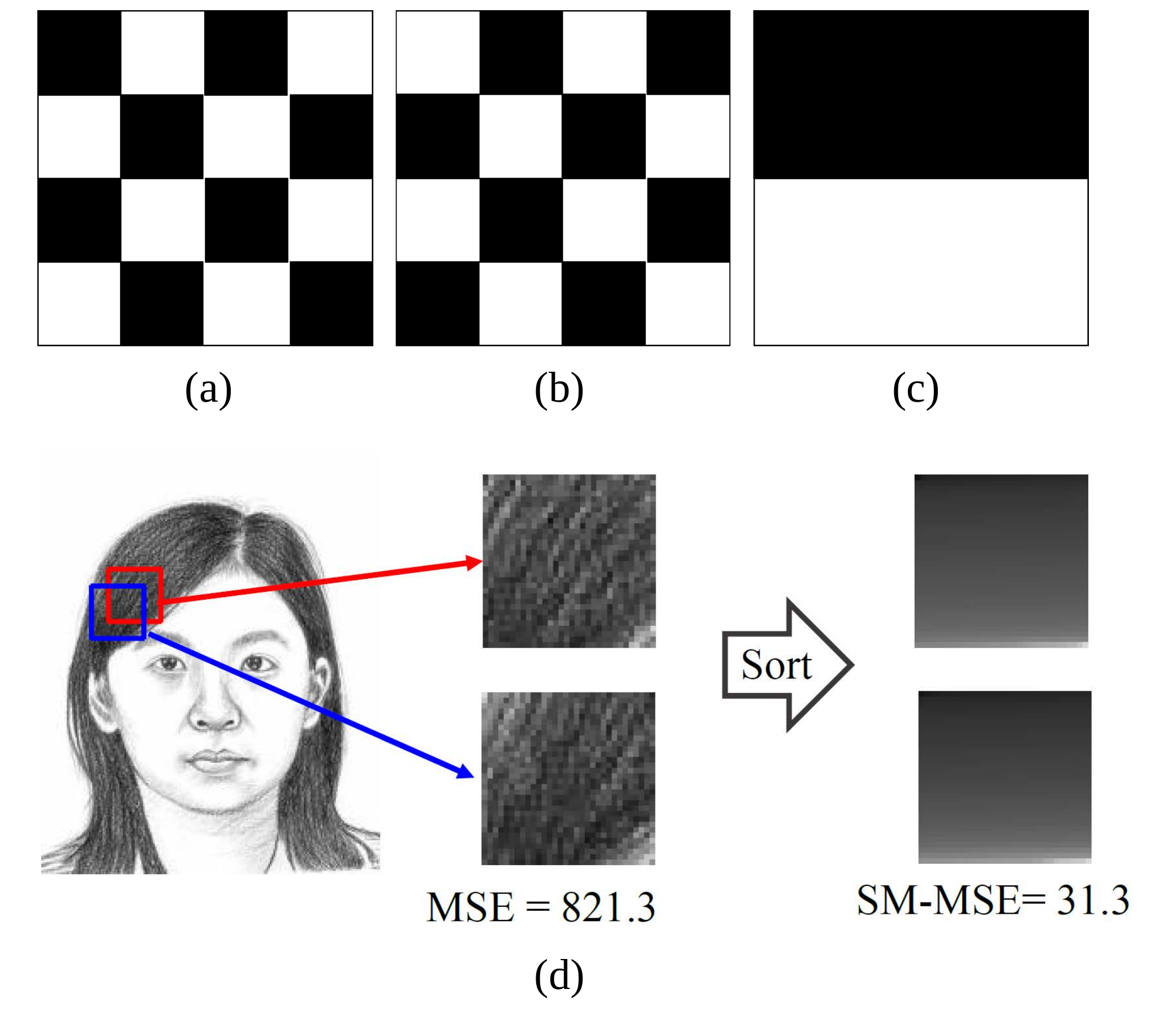}

\caption{Illustration of sorted matching. After applying the sort operator, two chessboard texture patterns in (a) and (b) become identical in (c); (d) Comparison of MSE and SM-MSE on textural pattern measurement}
\label{fig:illustraction-sm}
\end{figure}

Patches from hair regions are fed to BFCN for textural representation. Portrait textures usually contain fine-scale details with periodic and oscillatory natures. For example, the patches in Fig. \ref{fig:illustraction-sm} (a) and \ref{fig:illustraction-sm}(b) have visible point-by-point difference, but they are in the same texture pattern.
In this situation, directly applying a point-to-point objective function, e.g., mean square error (MSE), is difficult to evaluate the similarity of these similar textural patterns. Although extensive studies have been made on metrics of texture similarity \cite{zuo2013structral,chen2005adaptive,zujovic2013structural, wang2015kernel}, and many metrics has been proposed, they are difficult to be integrated into the neural network. For examples, the formulation of STSIM \cite{zujovic2013structural} is quite complex and hard to calculate the derivatives for back-propagation algorithm.

To deal with this situation, we design a \textit{Sorted Matching-Mean Square Error} (SM-MSE) metric for textural representation learning. SM-MSE can be viewed as adding an extra ascending sort operator before comparing two textural patches using MSE. We give an intuitive example of the comparison of adopting MSE and SM-MSE in Fig. \ref{fig:illustraction-sm}(d). We crop two close patches on the hair regions. Generally, those two patches are in the similar textural pattern. We apply the MSE and SM-MSE to evaluate the similarity of these patches respectively. As we can see, the result of SM-MSE is much smaller than those of directly applying MSE. Thus, by using SM-MSE, the similarity of two textural patches can be easily measured.
Besides, it is very straightforward to integrate SM-MSE into BFCN. We only need to mark down the index of each pixel before applying the sort operator, and then networks can find paths for back-propagating the derivatives, which is analogous to implement the back-propagation of the max pooling operator.

To train the textural branch of BFCN, we mainly adopt the combination of SM-MSE and MSE. Let $(p_t, s_t)$ denote a training patch pair for textural representation learning, $\mathbf{w}_g$ denote the parameters in shared layers and $\mathbf{w}_t$ denote parameters in the textural branch, respectively. Then the textural objective function $\mathcal{L}_{t}$ can be formulated as

\begin{equation}
\mathcal{L}_t = \frac{1}{|\mathbf{P}_t|}\sum_{p_t \in \mathbf{P}_t} \mathrm{MSE}(\hat{s}_t, s_t) + \beta \mathrm{SM}(\hat{s}_t, s_t),
\label{equation:textural-generation-objective}
\end{equation}
where $\hat{s}_{t}=f(p_t, \mathbf{w}_g, \mathbf{w}_{t})$ denotes the textural prediction of $s_t$, $\beta$ is used to balance the $\mathrm{MSE}(\cdot)$ and $ \mathrm{SM}(\cdot)$ term. The $\mathrm{MSE}(\cdot)$ term can be regarded as a regularizer.  Then, the $\mathrm{MSE}(\cdot)$ and $\mathrm{SM}(\cdot)$ in Eq. \ref{equation:textural-generation-objective} can be formulated as
\begin{equation}
\mathrm{MSE}(\hat{s}_t, s_t) = \frac{1}{|s_t|}\sum_{s_t^i \in s_t} \left ( \hat{s}_{t}^i - s_t^i \right ) ^ 2 ,
\label{equation:MSE2}
\end{equation}

\begin{equation}
\mathrm{SM}(\hat{s}_t, s_t) = \frac{1}{|s_{ts}|}\sum_{s_{ts}^i \in s_{ts}} \left ( \hat{s}_{ts}^i - s_{ts}^i \right ) ^ 2 ,
\label{equation:sort-MSE-3}
\end{equation}
where $s_t^i$ denotes the $i$-th ground truth pixel of a textural sketch patch $s_t$, and $\hat{s}_t^i\in\hat{s}_t$ denotes its prediction.
The $s_{ts}$ and $\hat{s}_{ts}=f_s(p_t, \mathbf{w}_g, \mathbf{w}_{t})$ are obtained by applying the ascending sort operator on $s_{t}$ and $\hat{s}_{t}$.
$s_{ts}^i$ denotes the $i$-th sorted ground truth pixel of a textural sketch patch $s_{ts}$, and $\hat{s}_{ts}^i\in\hat{s}_{ts}$ denotes the $i$-th sorted prediction.

\subsection{Probabilistic Fusion}

By using the parsing maps, we propose a probabilistic fusion scheme to fuse the structural and textural sketches directly to generate sketch portrait in the inference stage. The fusion process is guided by the probability parsing map of test photo $\mathbf{I}$ of size ${m\times n}$. Let $\mathbf{P}_f$, $\mathbf{P}_h$, $\mathbf{P}_b$ denote the probabilities of pixels in $\mathbf{I}$ belongs to face, hair and background respectively. We can obtain a binary map $\mathbf{P}_l$ which indicates whether pixels in $\mathbf{I}$ belongs to the hair or not, which can be formulated as
\begin{equation}
\mathbf{P}_l =  \mathbf{l}(\mathbf{P}_h  \geq \mathbf{P}_f \quad and \quad \mathbf{P}_h  \geq \mathbf{P}_b),
\end{equation}
where $\mathbf{l}(\cdot)$ denotes the indicator function. We then use $\mathbf{P}_l$ to fuse the structural sketch $\mathbf{S}_s$ and textural sketch $\mathbf{S}_t$ as
\begin{equation}
\mathbf{S} =  (\mathbf{1}_{m\times n}-\mathbf{P}_l) \cdot \mathbf{S}_s + \mathbf{P}_l \cdot \mathbf{S}_t.
\end{equation}
where $\mathbf{S}$ denotes the final sketch portrait.

\begin{figure}[thbp] \centering
\begin{minipage}[b]{0.5\textwidth} \centering
\includegraphics[width=0.8\textwidth]{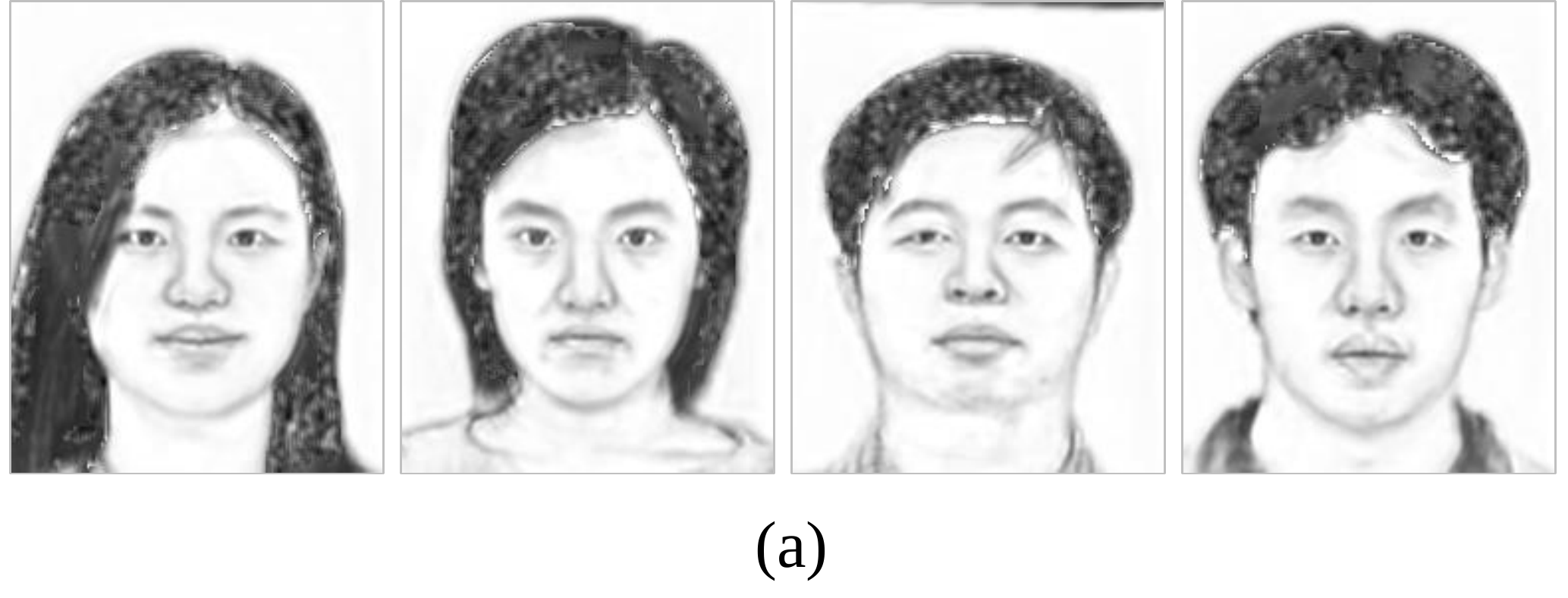}
\end{minipage}
\\\vspace*{0.1cm}
\begin{minipage}[b]{0.5\textwidth} \centering
\includegraphics[width=0.8\textwidth]{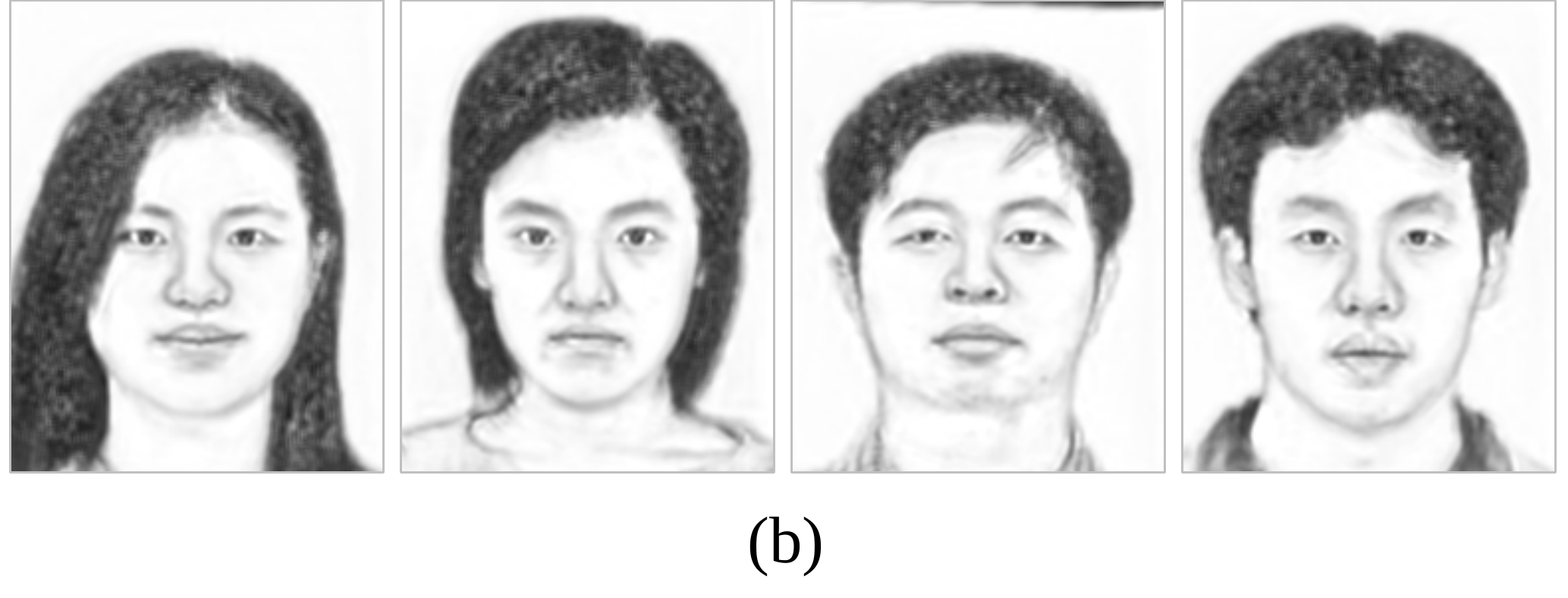}
\end{minipage}
\\\vspace*{0.1cm}

\caption{Comparison of different fusion strategies: (a) results of direct fusion, and (b) results of soft fusion.}
\label{fig:hard_soft_fusion}
\end{figure}

However, the above fusion process does not consider the border effect between the face and hair. Thus it may bring artifacts into final fusion results as shown in Fig. \ref{fig:hard_soft_fusion}(a). We can find sudden change between the border of face and hair. To overcome this problem, we propose a \textit{soft fusion} strategy. Instead of using the binary labels, the soft fusion adopt probability parsing maps to evaluate a weighted average between the structure-preserved sketch and texture-preserved sketch as:

\begin{equation}
\mathbf{S} = (\mathbf{1}_{m\times n} - \mathbf{P}_h) \cdot \mathbf{S}_s + \mathbf{P}_h \cdot \mathbf{S}_t,
\end{equation}
where $(\cdot)$ refers to element-wise product. By using soft fusion, the border between face and hair can be greatly smoothed.  A slice of samples of soft fusion are shown in Fig. \ref{fig:hard_soft_fusion}(b). Compared with Fig. \ref{fig:hard_soft_fusion}(a), we can see the border effects have been well removed.

\subsection{Implementation details}
We adopt the Caffe \cite{jia2014caffe} toolbox to implement both BFCN and P-Net.
For BFCN, the training samples are first cropped into size of $(156\times200)$ to exclude the influence of the black regions around the borders. Then, we crop the photo and its corresponding sketch into overlapping $(32 \times 32)$ patches to avoid overflow while keeping a high computational efficiency. In the training stage, filter weights of the two networks are initialized by drawing random numbers from a Gaussian distribution with zero mean and standard deviation 0.01, and the bias are initialized by zero.  We set $\alpha = 1$ and $\beta = 10$ for the hyper-parameters of the objective function in Eq. (\ref{equation:structrual-generation-objective}) and Eq. (\ref{equation:textural-generation-objective}). With the learning rate set as $10^{-10}$, BFCN needs about 150 epoches to converge. For the P-Net, it requires about 100 epoches to converge with learning rate $10^{-3}$.

In the inference stage, we adopt the $(200\times250)$ photos as input. In order to avoid the border effect, we do not use any paddings in the BFCN. Thus, the generated results will be shrunk to the size $(188\times238)$. Compared to most previous methods, our approach is very efficient (over 10 fps when processing aligned photos on a powerful GPU).

\section{Experimental Result}
In this section, we first introduce the datasets and implementation setting. Then, we conduct considerable experiments to show performance of our approach. The comparison results with some of existing methods are also discussed in this section.

\subsection{Dataset Setup}

\begin{figure}[thbp] \centering
\begin{minipage}[b]{0.50\textwidth} \centering
\includegraphics[width=0.14\textwidth]{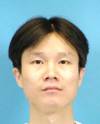}
\includegraphics[width=0.14\textwidth]{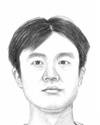}
\includegraphics[width=0.14\textwidth]{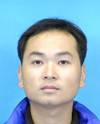}
\includegraphics[width=0.14\textwidth]{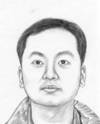}
\includegraphics[width=0.14\textwidth]{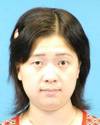}
\includegraphics[width=0.14\textwidth]{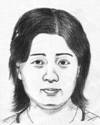}
\end{minipage}
\\\vspace*{0.1cm}
\begin{minipage}[b]{0.50\textwidth} \centering
\includegraphics[width=0.14\textwidth]{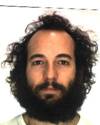}
\includegraphics[width=0.14\textwidth]{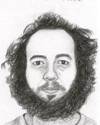}
\includegraphics[width=0.14\textwidth]{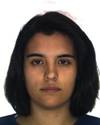}
\includegraphics[width=0.14\textwidth]{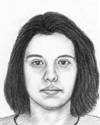}
\includegraphics[width=0.14\textwidth]{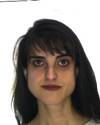}
\includegraphics[width=0.14\textwidth]{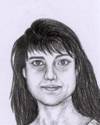}
\end{minipage}
\\\vspace*{0.1cm}
\begin{minipage}[b]{0.50\textwidth} \centering
\includegraphics[width=0.14\textwidth]{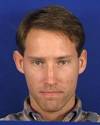}
\includegraphics[width=0.14\textwidth]{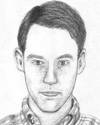}
\includegraphics[width=0.14\textwidth]{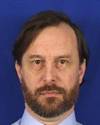}
\includegraphics[width=0.14\textwidth]{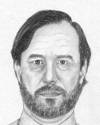}
\includegraphics[width=0.14\textwidth]{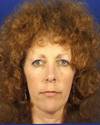}
\includegraphics[width=0.14\textwidth]{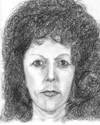}
\end{minipage}
\caption{Samples from the CUFS dataset. The samples are taken from the CUHK student dataset (the first row), the
AR dataset (the second row), and the XM2VTS dataset (the last row).}
\label{fig:testset-sample}
\end{figure}

For the sake of comparison with existing methods, we take the CUHK Face Sketch (CUFS) dataset \cite{wang2009face} for experimental study. The total samples of CUFS dataset is 606, which includes 188 samples from the Chinese University of Hong Kong (CUHK) student dataset, 123 samples from the AR dataset \cite{martinez1998ar}, and 295 samples from the XM2VTS dataset \cite{messer1999xm2vtsdb}. For each sample, there is a sketch drawn by an artist based on a photo taken in a frontal pose, under the normal lighting condition. Some samples from the CUFS dataset are shown in Fig. \ref{fig:testset-sample}. We take the 88 samples in CUHK student dataset as the training set, while the rest 518 samples are used as the testing set, including 123 samples from AR dataset, 295 samples from XM2VTS dataset and the reset 100 samples in CUHK student dataset.

We adopt the Helen dataset \cite{le2012interactive} and its additional annotations \cite{smith2013exemplar} to train the P-Net. We manually choose 280 samples in a roughly frontal pose assuming that the photos have been aligned by the landmarks.

\subsection{Photo-to-sketch Generation}
In this subsection, we evaluate the proposed framework on the CUFS dataset. We also compare our method with six recently proposed example-based synthesis methods, including Multiple Representations-based method (MR) \cite{PengMR2015}, Markov random field (MRF) \cite{wang2009face}, Markov weight field (MWF) \cite{zhou2012markov}, spatial sketch denoising (SSD) \cite{song_eccv14_sketch}, and sparse representation-based greedy search (SRGS) \cite{zhang2015face}.

\begin{figure*} [!thbp] \centering
\includegraphics[width=0.75\textwidth]{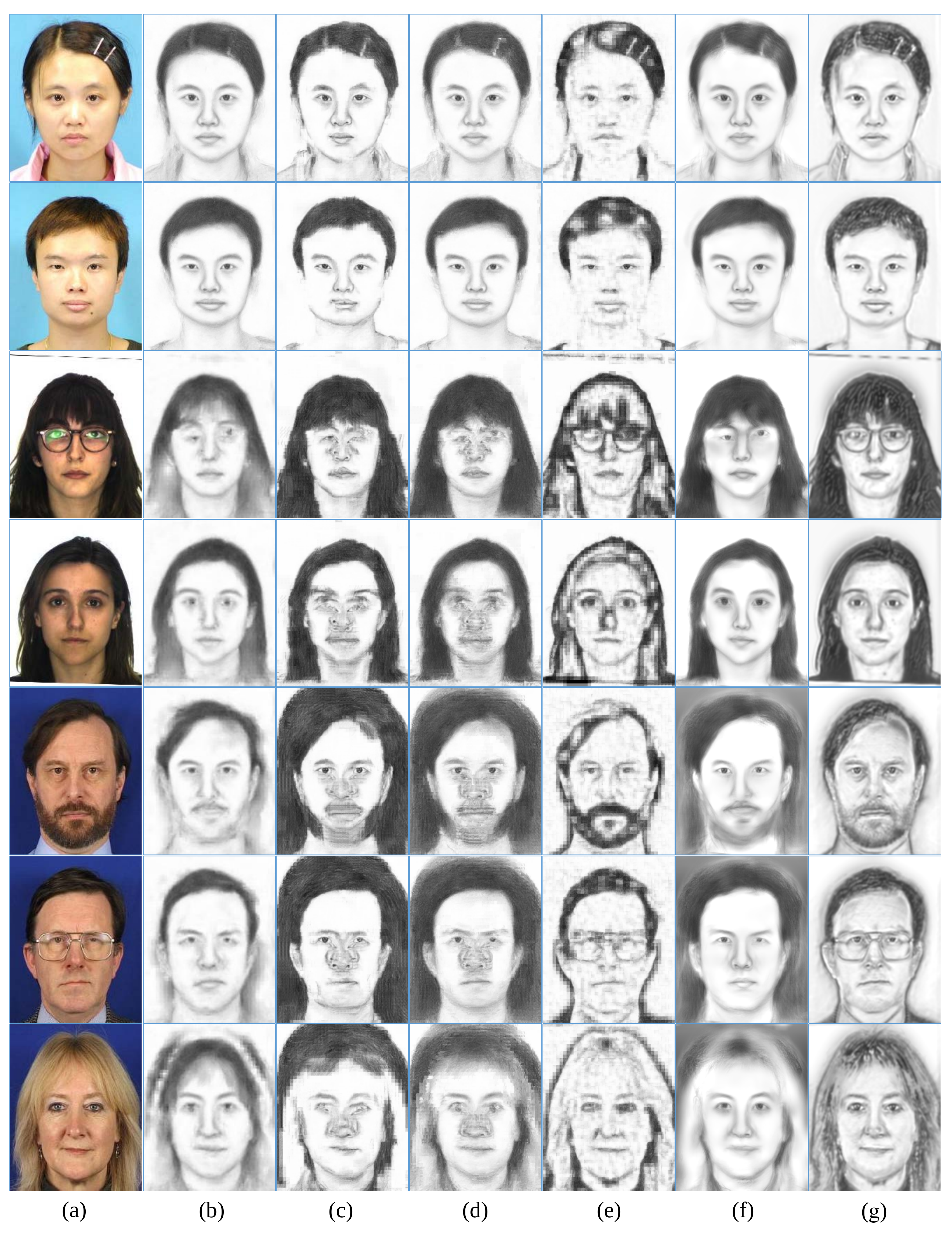}
\caption{Comparison of sketches generated by different methods. (a) Input Photo, (b) MR \cite{PengMR2015}, (c) MRF \cite{wang2009face}, (d) MWF \cite{zhou2012markov}, (e) SRGS \cite{zhang2015face}, (f) SSD \cite{song_eccv14_sketch}, (g) Our method.}
\label{fig:result-cufs}
\end{figure*}
The comparison results are shown in Fig. \ref{fig:result-cufs}. The first column corresponds to the input photos from CUHK, AR and XM2VTS, and the rest columns correspond to the sketches generated by MR \cite{PengMR2015}, MRF \cite{wang2009face}, MWF \cite{zhou2012markov}, SRGS \cite{zhang2015face}, SSD \cite{song_eccv14_sketch} and our method respectively. We can see that the visual effects of competing methods are not satisfactory. First, these methods can not handle decorations well, such as the hair pin in the first example and the glasses in the third and sixth examples. Besides, only our result exactly keeps the pigmented naevus in the input photo of the second row. Second, the competing methods can not preserve the fine-grained textural detail well. Especially when there are many texture regions in the sketch, e.g., the mustache and the hair regions. Compared with other methods, our approach can not only catch the significant characteristics of input photo portrait, but also preserve the fine-scale texture details to make the sketch portraits more vivid.

Another superiority of the proposed method is its generalization ability. In Fig. \ref{fig:result-cufs}, the results of the first two rows are more or less acceptable, while the rest results of other methods, i.e., images from the third row to the last row, are much worse in visual quality. This is because that the first two test photos are selected from CUHK student dataset, which shares the same distribution with the training samples, while the rest examples are taken from the AR and XM2VTS datasets, with different distributions from CUHK student dataset. Nevertheless, our method performs well on all input photos, showing its excellent generalization performance. 

Besides, the proposed decompositional representation learning based model can produce clearer structure and handle the non-facial factors better. For example, in Fig. \ref{fig:result-cufs}, the results produced by our method have clearer and sharper outliers of face, and preserve subtle structure of eyebrow, eyes, nose, lips, nose and ears. Take ears as example. The results generated by our method are satisfying, with fairly perfect shape and subtle detail preserved, while those produced by other methods are nearly unrecognizable. Meanwhile, only SRGS \cite{zhang2015face} and our methods can produce the non-facial factors, such as hairpin. However, SRGS  loses much fine-grained textural detail, such as the hair region of samples in Fig. \ref{fig:result-cufs}(e). In contrast, our method performs well in handling the fine-scale textural detail which makes our result much more vivid than those of SRGS.

%

\begin{figure} [thbp]
\centering
\includegraphics[width=0.45\textwidth]{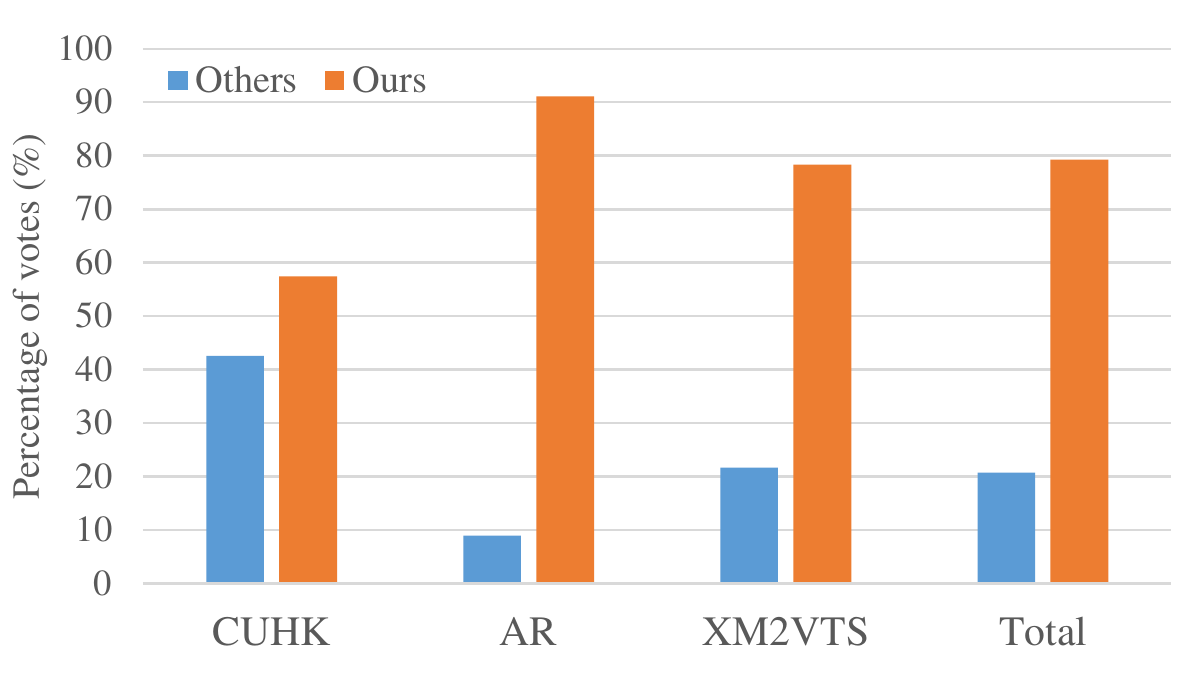}
\caption{Comparison on subjective voting. More people prefer the results generated by our approach.}
\label{fig:subjective-voting}
\end{figure}

Referring to \cite{song_eccv14_sketch, gao2012face}, we adopt subjective voting for the sketch image quality assessment. We present the candidate photos and the corresponding sketches produced by our method and other methods, including MR \cite{PengMR2015}, MRF \cite{wang2009face}, MWF \cite{zhou2012markov}, SSD \cite{song_eccv14_sketch} and SRGS \cite{zhang2015face}, and shuffle them.  We invited 20 volunteers to select the results that they prefer. The result is shown in Fig. \ref{fig:subjective-voting}, in which the blue bars refer to the percentage of votes selecting other methods, while the orange bars indicate the vote rate of our method. The statistic results show that much more people prefer our method. Specifically, for the CUHK dataset, our approach obtain over a half of all the votes. For other datasets, our superiority becomes more obvious, reaching 91\% and 78\% in AR and XM2VTS datasets, respectively.

\begin{figure}[thbp]\centering
\subfloat[Rank-1 Cumulative Match Score]{
\begin{minipage}[b]{0.5\textwidth}\centering
\includegraphics[width=0.75\textwidth]{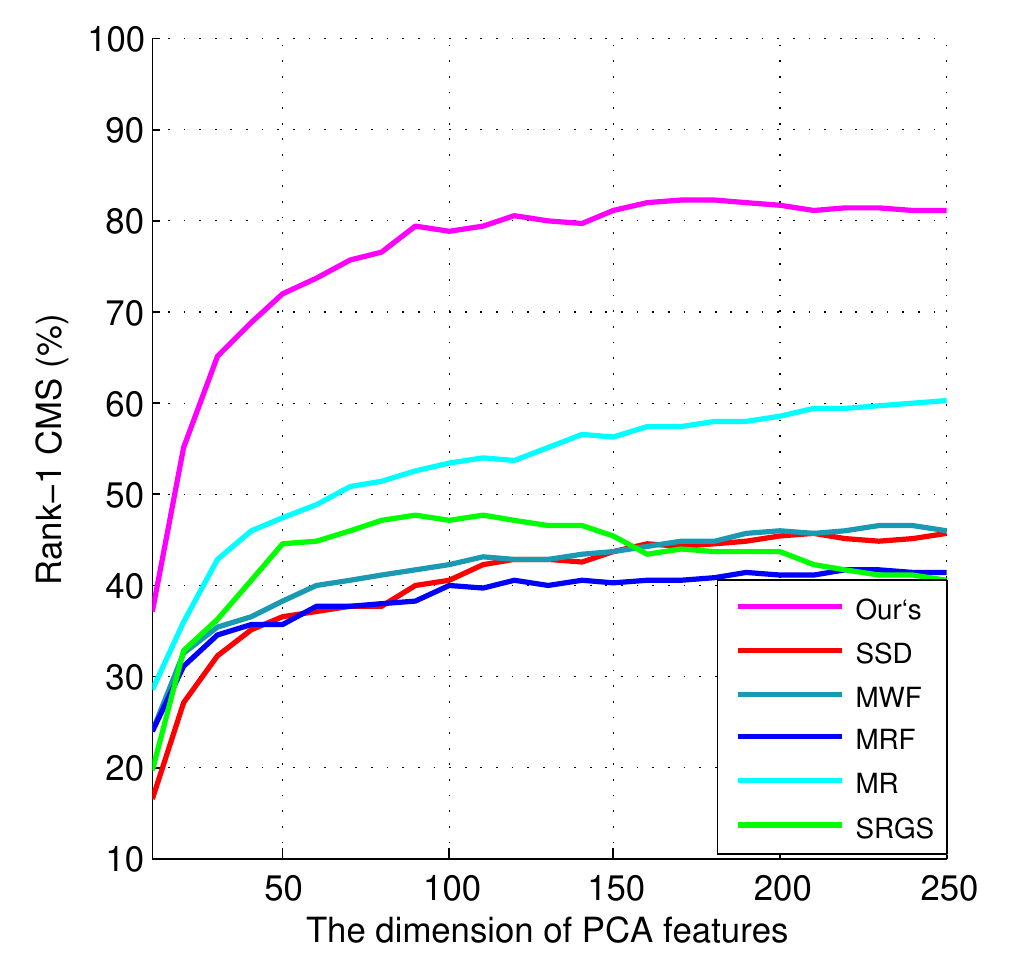}
\end{minipage}}\\

\subfloat[Rank-10 Cumulative Match Score]{
\begin{minipage}[b]{0.5\textwidth} \centering
\includegraphics[width=0.75\textwidth]{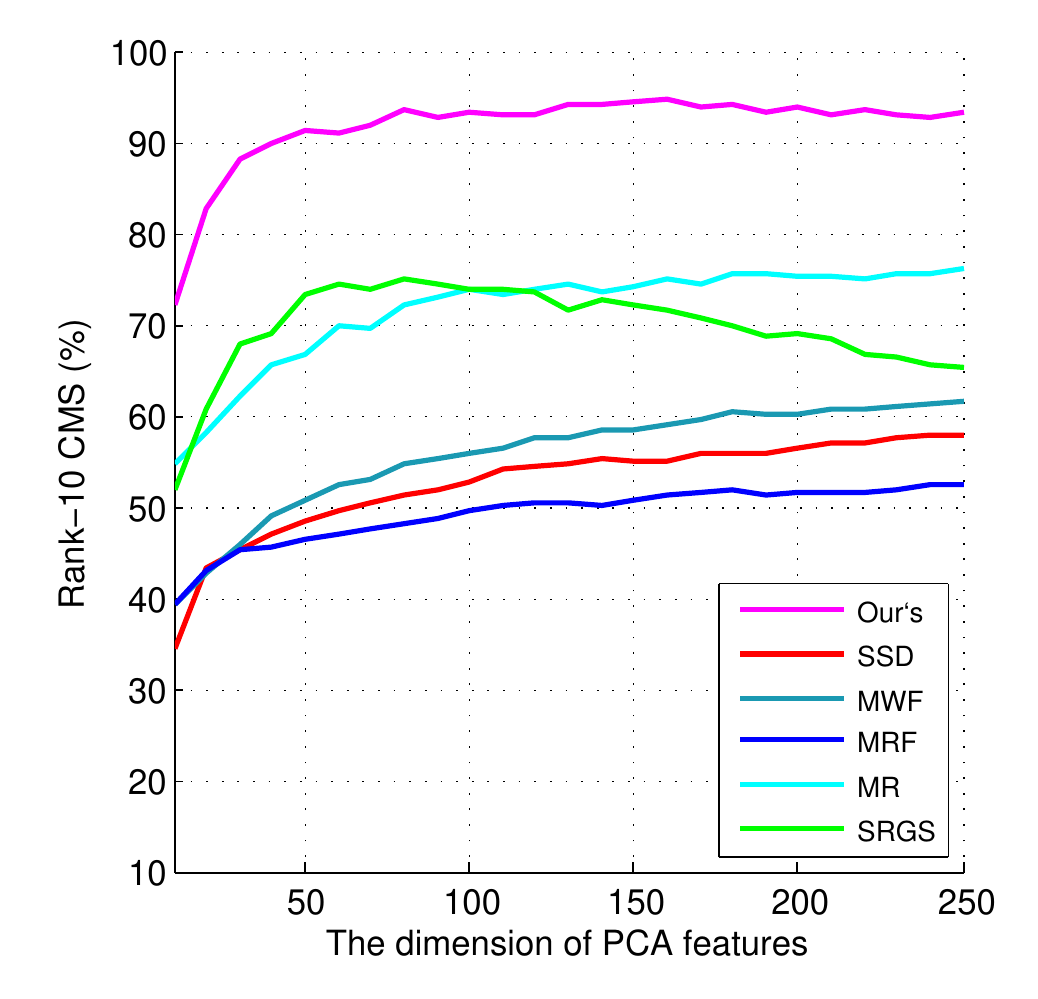}
\end{minipage}}

\caption{Comparison on the Rank-1 and Rank-10 Cumulative Match Score of sketch-based face recognition task. Best view in color.}
\label{fig:recognition-rankzzz}
\end{figure}

\subsection{Sketch-based Face Recognition}
The performance on sketch-based face recognition \cite{tang2004face} can also be used to evaluate the quality of sketch portraits.
In this subsection, we will show that the generated sketches of our proposed approach can not only get a high visual quality, but also can significantly reduce the modality difference between photos and sketches, which means our model can perform well on sketch-based face recognition task.

The procedures of a sketch-based face recognition can be concluded in two steps : (a) convert photos in testing set into corresponding sketches; (b) define a feature or transformation to measure the distance between the query sketch and the generated sketches.

We adopt PCA for face feature extraction and cosine similarity for distance measurement. Following the same protocol in \cite{tang2004face}, we compare our approach with previous methods on cumulative match score (CMS). The CMS measures the percentage of `the correct answer is in the top $n$ matches', where $n$ is called the rank.
We merge the total 518 samples from the CUHK, AR and XM2VTS datasets together to form a challenging sketch based recognition test set. In Fig. \ref{fig:recognition-rankzzz}(a), we plot the Rank-1 recognition rates of the comparison methods. The result of our method can get an accuracy of 78.7\% for the first match when using an 100-dimension PCA-reduced features, which is much better than the second place method (SRGS method \cite{zhang2015face}, 53.2\%). When the feature dimensions increase to 250, the Rank-1 CMS of our method also increases to 80.1\%. As shown in Fig. \ref{fig:recognition-rankzzz}(b), our method can reach to a accuracy of 93.2\% in ten guesses, while the best result of other methods is around 85\%.

\subsection{Robustness to Lighting and Pose Variations}
\begin{figure}[t]
\centering
\subfloat[Photos]{
\begin{minipage}[b]{0.08\textwidth}
\includegraphics[width=1\columnwidth]{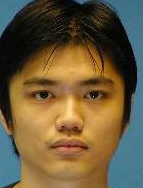}
\\\vspace*{-0.3cm}
\includegraphics[width=1\columnwidth]{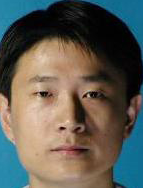}
\\\vspace*{-0.3cm}
\includegraphics[width=1\columnwidth]{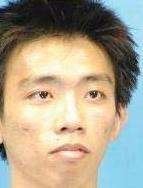}
\\\vspace*{-0.3cm}
\includegraphics[width=1\columnwidth]{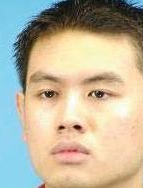}
\end{minipage}
}\hspace*{-0.2cm}
\subfloat[MRF]{
\begin{minipage}[b]{0.08\textwidth}
\includegraphics[width=1\columnwidth]{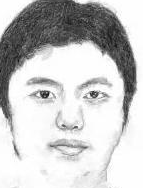}
\\\vspace*{-0.3cm}
\includegraphics[width=1\columnwidth]{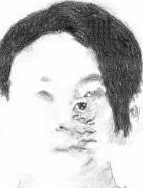}
\\\vspace*{-0.3cm}
\includegraphics[width=1\columnwidth]{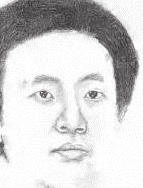}
\\\vspace*{-0.3cm}
\includegraphics[width=1\columnwidth]{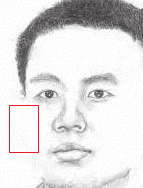}
\end{minipage}
}\hspace*{-0.2cm}
\subfloat[MRF+]{
\begin{minipage}[b]{0.08\textwidth}
\includegraphics[width=1\columnwidth]{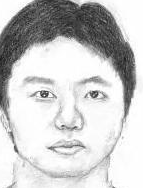}
\\\vspace*{-0.3cm}
\includegraphics[width=1\columnwidth]{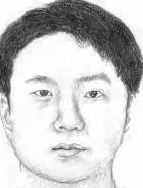}
\\\vspace*{-0.3cm}
\includegraphics[width=1\columnwidth]{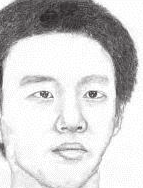}
\\\vspace*{-0.3cm}
\includegraphics[width=1\columnwidth]{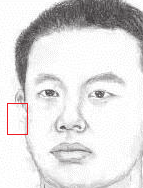}
\end{minipage}
}\hspace*{-0.2cm}
\subfloat[Ours]{
\begin{minipage}[b]{0.08\textwidth}
\includegraphics[width=1\columnwidth]{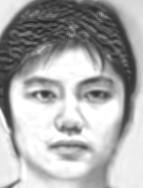}
\\\vspace*{-0.3cm}
\includegraphics[width=1\columnwidth]{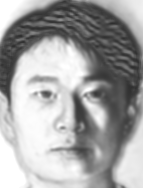}
\\\vspace*{-0.3cm}
\includegraphics[width=1\columnwidth]{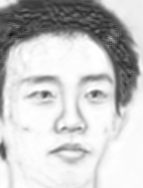}
\\\vspace*{-0.3cm}
\includegraphics[width=1\columnwidth]{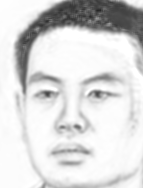}
\end{minipage}
}\hspace*{-0.2cm}
\subfloat[Ours+]{
\begin{minipage}[b]{0.08\textwidth}
\includegraphics[width=1\columnwidth]{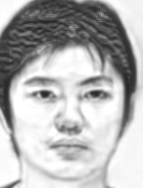}
\\\vspace*{-0.3cm}
\includegraphics[width=1\columnwidth]{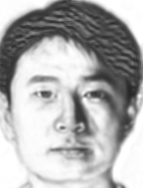}
\\\vspace*{-0.3cm}
\includegraphics[width=1\columnwidth]{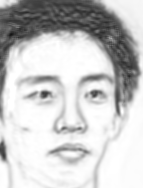}
\\\vspace*{-0.3cm}
\includegraphics[width=1\columnwidth]{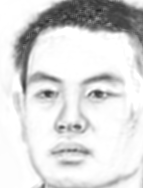}
\end{minipage}
}\hspace*{-0.2cm}
\caption{Comparison of the robustness to lighting and pose variations of different methods.}
\label{fig:result-variation}
\end{figure}

The lighting and pose variations are also challenging in the sketch generation problem \cite{zhang2010lighting}. Some of previous methods only work under well constrained conditions and often fail when there are variations of lighting and pose. For example, Fig. \ref{fig:result-variation}(b) shows the samples of sketches synthesized by MRF \cite{wang2009face} methods with lighting and pose variations. The results of the first and second rows are obtained under dark frontal lighting and dark side lighting, while the results of the third and fourth rows are synthesized under pose variations in range of $[-45^\circ, 45^\circ]$. The results show that MRF often lose some details under lighting and pose variations. For example, in the sketch of the forth row of Fig. \ref{fig:result-variation}(b), the profile and ear is missed, and the sketch in the second row is dramatically confused.  Zhang et al. \cite{zhang2010lighting} further improved MRF (named as MRF+ in this paper) to handle the lighting and pose variations. However, MRF+ involves much additional operations which make it rather complicated and inefficient. The results of the MRF+ are shown in Fig. \ref{fig:result-variation}(c). We can see that the visual effect of the MRF+ is improved, however, the results still lack some details, e.g., part of the ear marked in the forth row of Fig. \ref{fig:result-variation}(c).

Our proposed method learns the sketch from the raw pixels of photo portrait, and it is rather robust to the pose variation as shown in the third and forth row of Fig. \ref{fig:result-variation}(d) and (e). Besides, we can adopt a simple strategy to handle the lighting variation. Specifically, we first translate the input photos to HSV colors pace, and then randomly multiple the index of V channel by a factor in the range $[0.625, 1.125]$ during the training. The sketch results are shown in the first and second row of Fig. \ref{fig:result-variation}(e). Compared with the corresponding sketches of Fig. \ref{fig:result-variation}(d) , the visual effects are marginally improved.

\subsection{Portrait-to-sketch Generation in the Wild}
\begin{figure*}[thbp]\centering
\includegraphics[width=0.75\textwidth]{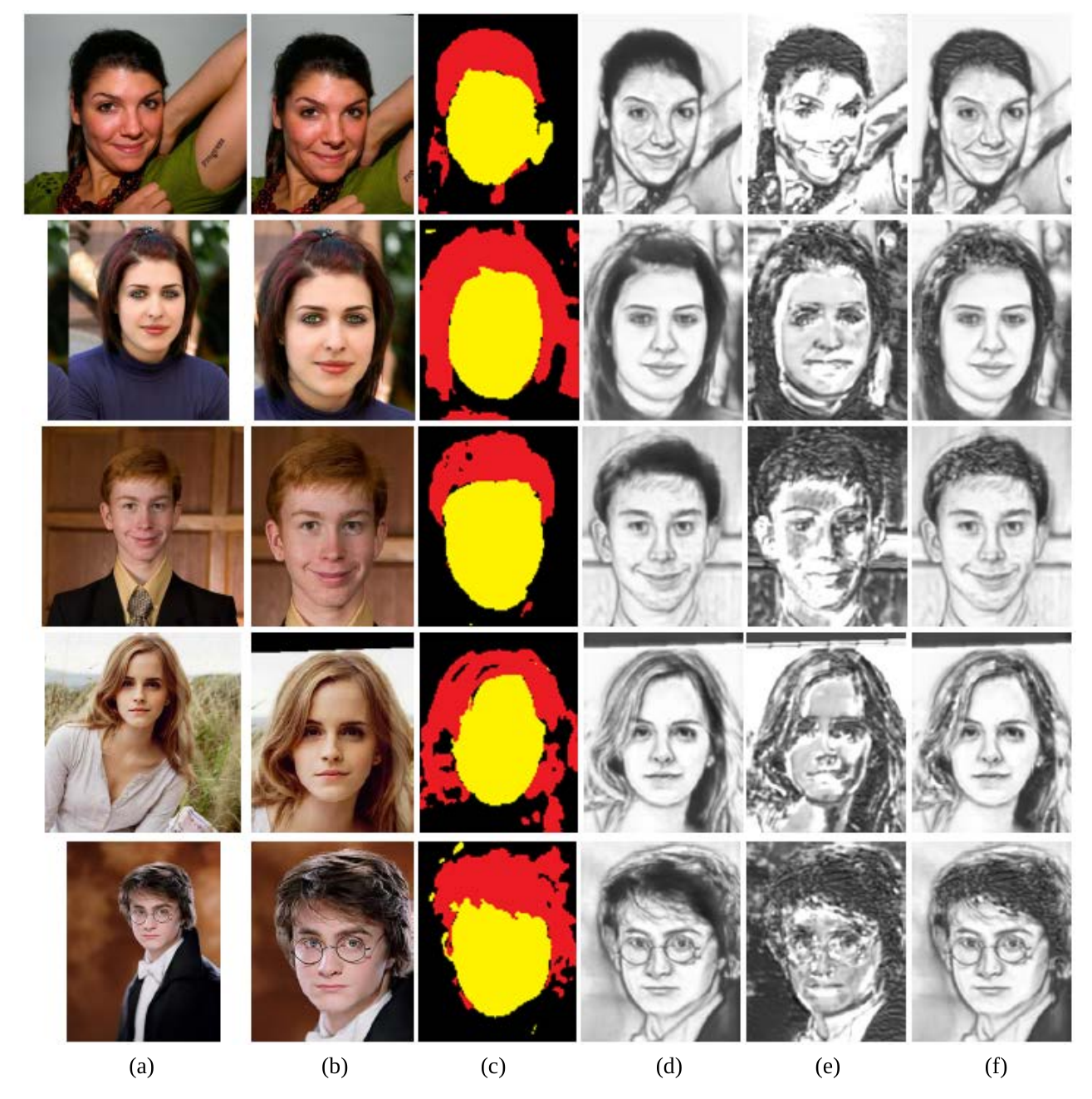}
\caption{Results generated by our framework in unconstrained environment. (a) Input portraits; (b) aligned portraits; (c) parsing map; (d) structural sketches; (e) textural sketches; (f) fused sketches.}
\label{fig:sample-in-the-wild}
\end{figure*}

In this section, we conduct experiments to explore generation ability of our model on an unconstrained environment.
We select some generated sketch portraits and show them in Fig. \ref{fig:sample-in-the-wild}  with corresponding intermediate results. It indicates that the representation learned by our model is more general and more robust to handle the complex background (e.g., the left arm of the woman in the first row, and the batten behind the man in the third row).

\subsection{Analysis and Discussion}
We also analysis the effectiveness of the decompositional representation learning and parsing maps in the proposed method. Besides, we also discuss some considerations when designing the probabilistic fusion and the architecture of BFCN.

\subsubsection{The effectiveness of decompositional representation learning}
\begin{figure}[t]
\centering
\subfloat[]{
\begin{minipage}[b]{0.1\textwidth}
\includegraphics[width=1\columnwidth]{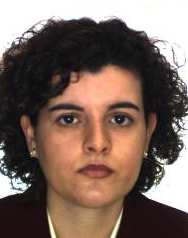}
\\\vspace*{-0.3cm}
\includegraphics[width=1\columnwidth]{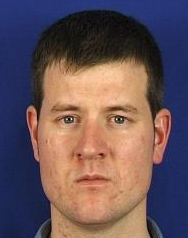}
\end{minipage}
}\hspace*{-0.2cm}
\subfloat[]{
\begin{minipage}[b]{0.1\textwidth}
\includegraphics[width=1\columnwidth]{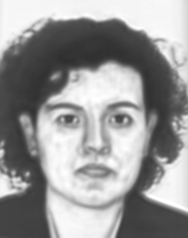}
\\\vspace*{-0.3cm}
\includegraphics[width=1\columnwidth]{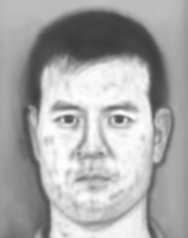}
\end{minipage}
}\hspace*{-0.2cm}
\subfloat[]{
\begin{minipage}[b]{0.1\textwidth}
\includegraphics[width=1\columnwidth]{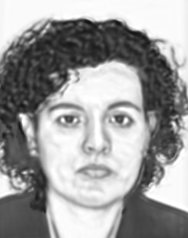}
\\\vspace*{-0.3cm}
\includegraphics[width=1\columnwidth]{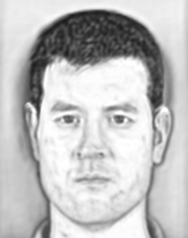}
\end{minipage}
}\hspace*{-0.2cm}
\caption{Comparison on models trained without/with decompositional representation learning (DRL). (a) Input photos; (b) Results without DRL; (c) Results with DRL. }
\label{fig:drl-vs-nodrl}
\end{figure}

We conduct experiments to verify the effectiveness of decompositional representation learning on handling the structures and textures. Specifically, we disable the structurally unaligned filter in the data preparing stage, and set $\beta = 0$ to remove $SM(\cdot)$ term in Eq. (\ref{equation:textural-generation-objective}) when training the BFCN. Under this setting, the two branches of BFCN are trained equally with the same loss function. Then we retrain the model under this condition. The results are depicted in the second column of Fig. \ref{fig:drl-vs-nodrl}. For comparison, we also depict the result with normal setting in the third column. Obviously, the sketches in the third column are more attractive. The textures are much clearer, since SM-MSE metric can correctly evaluate similar textures to learn a better representation. Meanwhile, the structures are sharper, since the structurally unaligned filter only retains the aligned patch pairs, which help to capture the main structures and suppress the noises.


\begin{figure}[t]
\centering
\subfloat[]{
\begin{minipage}[b]{0.1\textwidth}
\includegraphics[width=1\columnwidth]{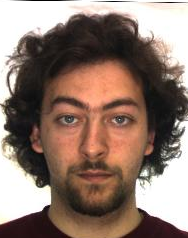}
\\\vspace*{-0.3cm}
\includegraphics[width=1\columnwidth]{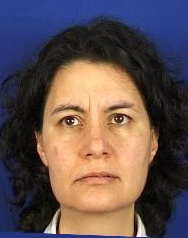}
\end{minipage}
}\hspace*{-0.2cm}
\subfloat[]{
\begin{minipage}[b]{0.1\textwidth}
\includegraphics[width=1\columnwidth]{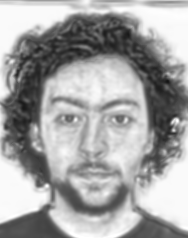}
\\\vspace*{-0.3cm}
\includegraphics[width=1\columnwidth]{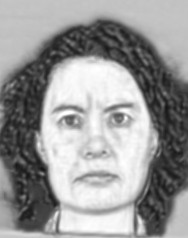}
\end{minipage}
}\hspace*{-0.2cm}
\subfloat[]{
\begin{minipage}[b]{0.1\textwidth}
\includegraphics[width=1\columnwidth]{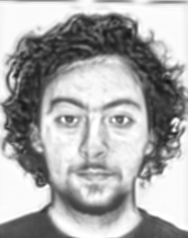}
\\\vspace*{-0.3cm}
\includegraphics[width=1\columnwidth]{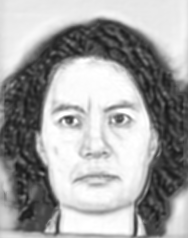}
\end{minipage}
}\hspace*{-0.2cm}
\caption{Comparison results of model trained without/with the nonparametric prior. (a) Input photos; (b) Results without global prior; (c) Results with global prior. }
\label{fig:prior-vs-noprior}
\end{figure}

\subsubsection{The effectiveness of nonparametric prior in training BFCN} As we mentioned in Section III, in the training of BFCN, we add the average of ground truth of sketch as nonparametric prior to provide a global regularization to our model. Here, we evaluate the role of this nonparametric prior via comparing the sketches generated by the models with and without this prior respectively. The comparison results are presented in Fig. \ref{fig:prior-vs-noprior}. We can see that after embedding the nonparametric prior into our model, some mistakes caused by the locally predictions are corrected and the sketches are more lively.

\subsubsection{Shared vs. unshared parameters of shallow layers}
The low-level feature learned by SRCNN \cite{dong2014learn} is likely to be edges, which can be shared in most of the computer vision tasks. Inspired by previous works \cite{dong2014learn, girshick2015fast}, we share parameters of the first three convolutional layers (called shallow layers) of BFCN and we find that this strategy is both effective and efficient.
For comparison, we retrain a model without sharing the parameters, i.e., we adopt two isolated networks to learn the structures and textures. Experimental results show that sharing the shallow layers is much more efficient. As shown in TABLE \ref{table:share-layer-times}, if we don't share the weights, testing procedure will be significantly slowed down by over 110\%, since most of the computational cost comes from the shallow convolutional layers. Besides, we also compared the computation cost of proposed BFCN with other methods, i.e., MRF \cite{wang2009face}, SSD \cite{song_eccv14_sketch}, SRGS \cite{zhang2015face}, MR \cite{PengMR2015}, MWF \cite{zhou2012markov} to evaluate its efficiency. For fair comparison, all of these methods are run on a PC with Intel Core i7 3.4GHz CPU without GPU acceleration.  The comparison results are list in Table \ref{table:time_comparison} show that our method is much more efficient than other methods.

\begin{table}[thbp] \centering \footnotesize
\center
\caption{Inference time for single image of unshared and shared parameters of shallow layers (On a NVIDIA Titan Black GPU).}
\begin{tabular}{*{3}{c}}
\toprule
 & Unshared & Shared \\
\midrule
Time(ms) & 63.0  & \textbf{29.8}  \\
\bottomrule
\end{tabular}

\label{table:share-layer-times}
\end{table}

\begin{table}[!thbp]\centering \footnotesize

\center
\caption{Comparison of inference time of single face image of different methods.}

\begin{tabularx}{0.5\textwidth}{p{0.6cm}*{6}{c}{p{0.6cm}}}
\toprule
&MRF\cite{wang2009face}&SSD\cite{song_eccv14_sketch}&SRGS\cite{zhang2015face}&MR\cite{PengMR2015}&MWF\cite{zhou2012markov}&Our\\
\midrule
Time(s)&155&4&4&600&40&\textbf{1.2} \\
\bottomrule
\end{tabularx}

\label{table:time_comparison}
\end{table}

\section{Conclusion and future work}
In this paper, we propose a novel decompositional representation learning framework to learn from the raw pixels of input photo for an end-to-end sketch portrait generation. We utilize a BFCN to map the photo into structural and textural components to generate a structure-preserved sketch and a texture-preserved sketch respectively. The two sketches are fused together to generate the final sketch portrait via a probabilistic method. Experimental results on several challenging benchmarks show the proposed method outperforms existing example-based synthesis algorithms in terms of both perceptual and objective metrics. Besides, the proposed approach also has favorable generalization ability across different datasets without additional training.

Currently, in the training BFCN, a face image and its corresponding sketch are roughly aligned by eyes. Then patches of face image and its corresponding sketch patches are fed into BFCN to train a photo-sketch generation model. In other words, the performance of BFCN is partially rely on the face alignment algorithm. If the face images have large pose variations or drastic lighting change, the results of current face alignment method may be not good. Thus the sketches generated by BFCN may be not satisfied. In the future, we will design more robust face alignment algorithm to replace current strategy, and make the BFCN more robust to the pose and lighting variations.

%



\ifCLASSOPTIONcaptionsoff
  \newpage
\fi

\bibliographystyle{IEEEtran}
\bibliography{IEEEabrv,sketch}


%
\vspace{-35pt}
\begin{IEEEbiography}[{\includegraphics[width=1in,height=1.25in,clip]{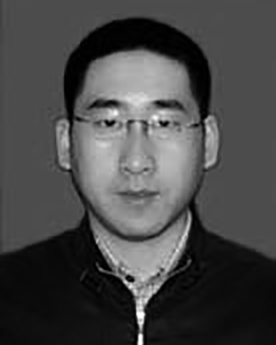}}]{Dongyu Zhang}
received the B.S. and Ph.D. degrees from the Harbin Institute of Technology, Harbin,China, in 2003 and 2010, respectively.
He is currently a Research Associate Professor with the School of Data and Computer Science, Sun Yat-sen University, Guangzhou, China. His current research interests include computer vision and machine learning.

\end{IEEEbiography}
\vspace{-25pt}

\begin{IEEEbiography}[{\includegraphics[width=1in,height=1.25in,clip]{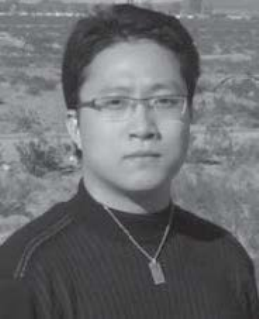}}]{Liang Lin} (SM'15) is a Professor with the School of Data and Computer Science, Sun Yat-sen University (SYSU), China. He received the B.S. and Ph.D. degrees from the Beijing Institute of Technology, Beijing, China, in 1999 and 2008, respectively. He was a Post-Doctoral Research Fellow with the Department of Statistics, University of California, Los Angeles. Dr. Lin’s research focuses on new models, algorithms and systems for intelligent processing and understanding of visual data. He has authorized or co-authorized more than 80 papers in top tier academic journals and conferences. He has served as an associate editor for IEEE Trans. Human-Machine Systems, Neurocomputing and The Visual Computer, and a guest editor for Pattern Recognition. He was supported by several promotive programs or funds for his works, such as the NSFC for Excellent Young Scientist in 2016. He received the Best Paper Runners-Up Award in ACM NPAR 2010, Google Faculty Award in 2012, and Best Student Paper Award in IEEE ICME 2014.

\end{IEEEbiography}

\vspace{-35pt}

\begin{IEEEbiography}[{\includegraphics[width=1in,height=1.25in,clip,keepaspectratio]{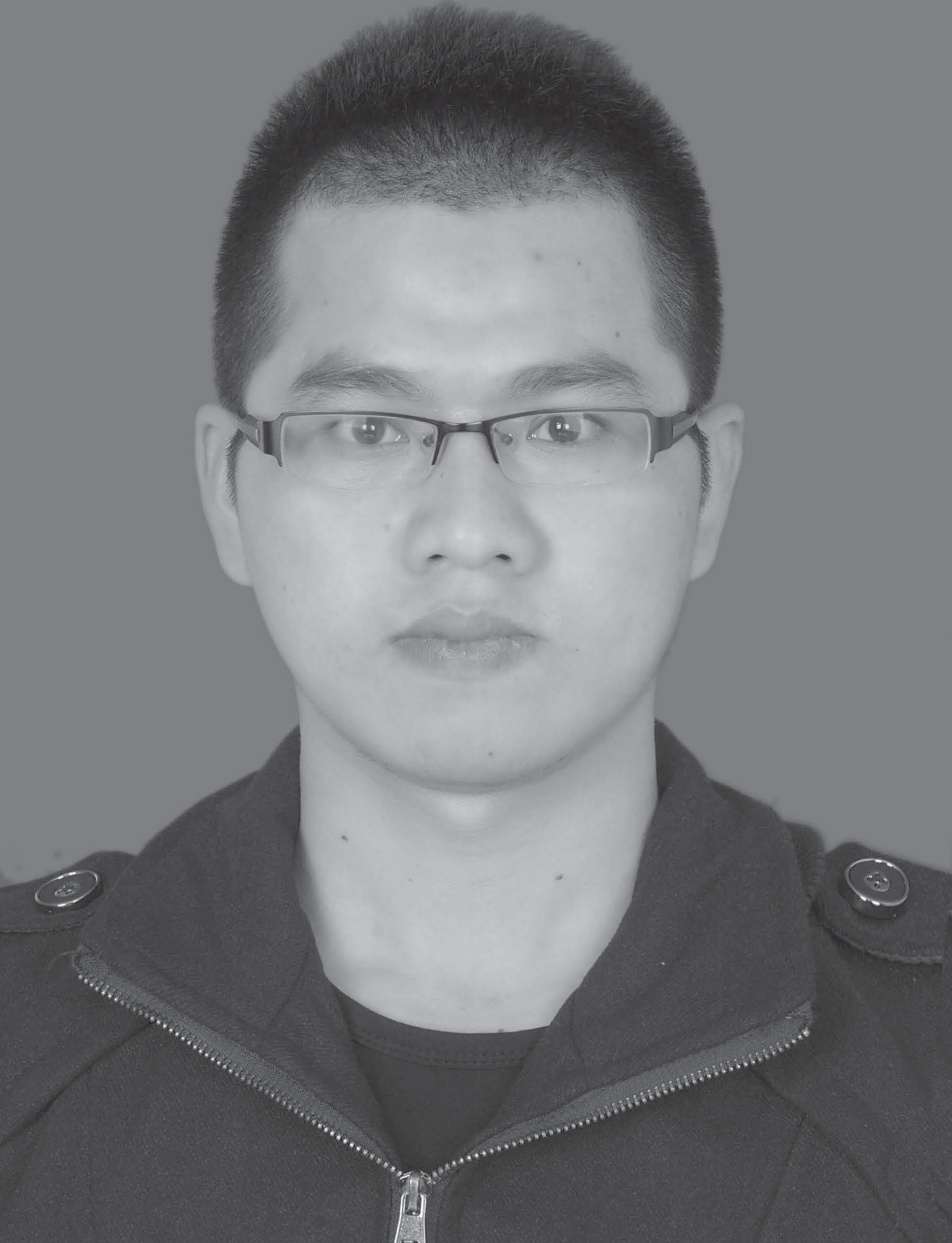}}]{Tianshui Chen}

received the B.E. degree from School of Information and Science Technology, Sun Yat-sen University, Guangzhou, China, in 2013, where he is currently pursuing the Ph.D. degree in computer science with the School of Data and Computer Science. His current research interests include computer vision and machine learning.

\end{IEEEbiography}
\vspace{-35pt}
\begin{IEEEbiography}[{\includegraphics[width=1in,height=1.25in,clip,keepaspectratio]{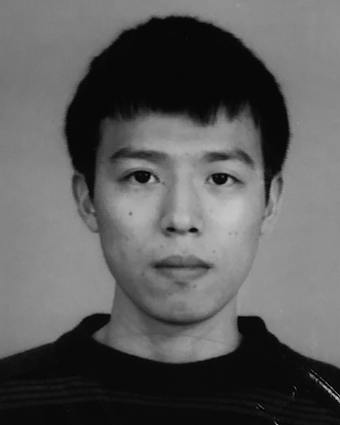}}]{Xian Wu}

received the B.E. degree from the School of Software, Sun Yat-sen University, Guangzhou, China, in 2015. He is currently pursuing the MS c. degree in Software Engineering with the School of Data and Computer Science. His current research interests include computer vision and machine learning.

\end{IEEEbiography}
\vspace{-35pt}

\begin{IEEEbiography}[{\includegraphics[width=1in,height=1.25in,clip,keepaspectratio]{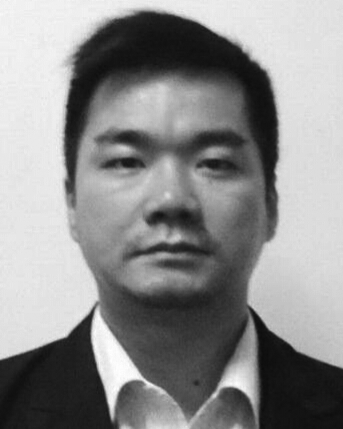}}]{Wenwei Tan}
is a senior research engineer in Hisilicon Technologies co., LTD. He received the B.S. and M.D. degrees from Guangdong University of Technology, Guangzhou, China, in 2005 and 2008, respectively. His currently research interests include deep learning chip and artificial intelligence.
\end{IEEEbiography}
\vspace{-35pt}

\begin{IEEEbiography}[{\includegraphics[width=1in,height=1.25in,clip]{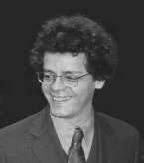}}] {Ebroul Izquierdo} (SM’03) received the M.Sc., C.Eng., Ph.D., and Dr. Rerum Naturalium (Ph.D.) degrees from Humboldt University, Berlin, Germany. He is currently the Chair of the Multimedia and Computer Vision, and the Head of the Multimedia and Vision Group with the School of Electronic Engineering and Computer Science, Queen Mary University of London, London, U.K. He has been a Senior Researcher with Heinrich Hertz Institute for Communication Technology, Berlin, and the Department of Electronic Systems Engineering, University of Essex, Colchester, U.K. He has authored over 450 technical papers, including chapters in books, and holds several patents in multimedia signal processing. Prof. Izquierdo is a Chartered Engineer, a Fellow Member of the Institution of Engineering and Technology (IET), a member of the British Machine Vision Association, the Chairman of the IET Professional Network on Information Engineering, a member of the Visual Signal Processing and Communication Technical Committee of the IEEE Circuits and Systems Society, and a member of the Multimedia Signal Processing Technical Committee of the IEEE. He has been an Associated and Guest Editor of several relevant journals, including IEEE TRANSACTIONS ON CIRCUITS AND SYSTEMS FOR VIDEO TECHNOLOGY, EURASIP Journal on Image and Video processing, Signal Processing: Image Communication (Elsevier), EURASIP Journal on Applied Signal Processing, IEEE PROCEEDINGS ON VISION, IMAGE AND SIGNAL PROCESSING, Journal of Multimedia Tools and Applications, and Journal of Multimedia.
\end{IEEEbiography}

\end{document}